\documentclass[]{spie}  

\usepackage{amsmath,amsfonts,amssymb}
\usepackage{graphicx}
\usepackage[colorlinks=true, allcolors=blue]{hyperref}
\usepackage[square, numbers]{natbib}
\usepackage{algorithm}
\usepackage{algpseudocode}
\usepackage{enumitem}

\title{Adaptive Augmentation Policy Optimization with LLM Feedback}

\author[]{Ant Duru}
\author[]{Alptekin Temizel}
\affil[]{Graduate School of Informatics, METU, Address, Ankara, Turkey}

\usepackage[utf8]{inputenc}  
\usepackage[T1]{fontenc}   
\begin{document} 
\maketitle

\begin{abstract}
Data augmentation is a critical component of deep learning pipelines, enhancing model generalization by increasing dataset diversity. Traditional augmentation strategies rely on manually designed transformations, stochastic sampling, or automated search-based approaches. Although automated methods improve performance, they often require extensive computational resources and are specifically designed for certain datasets. In this work, we propose a Large Language Model (LLM)-guided augmentation optimization strategy that refines augmentation policies based on model performance feedback. We propose two approaches: (1) LLM-Guided Augmentation Policy Optimization, where augmentation policies selected by LLM are refined iteratively across training cycles, and (2) Adaptive LLM-Guided Augmentation Policy Optimization, which adjusts policies at each iteration based on performance metrics. This in-training approach eliminates the need for full model retraining before getting LLM feedback, reducing computational costs while increasing performance. Our methodology employs an LLM to dynamically select augmentation transformations based on dataset characteristics, model architecture, and prior training performance. Leveraging LLMs’ contextual knowledge, especially in domain-specific tasks like medical imaging, our method selects augmentations tailored to dataset characteristics and model performance. Experiments across domain-specific image classification datasets show consistent accuracy improvements over traditional methods. The code for the adaptive approach can be found at \href{https://github.com/antduru/adaptive-augmentation-using-llms}{GitHub}.
\end{abstract}


\section{INTRODUCTION}
\label{sec:intro}  

Deep learning models have achieved remarkable success in various image classification tasks, but they often require large labeled datasets for robust generalization. In specialized fields, such as medical imaging, acquiring extensive labeled datasets is challenging due to data scarcity, ethical concerns, and high annotation costs. To address this limitation, data augmentation has become an essential technique, enhancing model performance by increasing the diversity of training data through transformations such as rotation, flipping, cropping, and contrast adjustments. The choice of augmentation strategy is critical, as suboptimal augmentations may fail to improve generalization or even degrade model performance. 

Traditional augmentation relies on manually designed transformations, requiring domain expertise to tailor policies to specific datasets. However, this approach is time-consuming and does not scale across different tasks. As the complexity of deep learning applications grows, manual augmentation tuning becomes impractical, necessitating the development of systematic, automated augmentation strategies. Search-based optimization techniques such as AutoAugment \cite{cubuk2019autoaugment}, RandAugment \cite{cubuk2019randaugment}, and TrivialAugment \cite{trivialaugment} systematically explore augmentation search spaces to identify policies that maximize model performance. While effective, they often require extensive computational resources and are sensitive to dataset characteristics, limiting their applicability to various real-world scenarios. Additionally, the augmentation strategies derived from these methods frequently lack flexibility, as they rely on pre-computed augmentation configurations that may not be optimal across different training stages or evolving model architectures.

Recently, large language models (LLMs) have emerged as powerful tools capable of optimizing various aspects of deep learning workflows, including hyperparameter tuning and automated machine learning. Prior studies have shown that LLMs can effectively suggest hyperparameters such as learning rates, batch sizes, and optimizers based on model performance feedback \cite{zhang2023llm}. Their capacity to process large amounts of unstructured knowledge and synthesize contextual insights makes them particularly well-suited for data-driven optimization tasks. Inspired by these advancements, we explore the use of LLMs to optimize data augmentation strategies for image classification. Unlike conventional search-based methods that lack semantic understanding, large language models (LLMs) are pretrained on extensive scientific and domain-specific literature. This enables them to guide effective augmentation policy selection in task-specific contexts. Furthermore, LLMs can incorporate high-level cues such as dataset modality, model architecture, or class distribution to suggest transformations that are more aligned with the underlying data structure. Their iterative feedback loop mimics an adaptive optimization process, enabling augmentation decisions to evolve as the model learns.

In this paper, we propose two automated augmentation strategies that use LLMs to generate dataset-specific augmentation policies. The main contributions can be summarized as follows:
\begin{itemize}[left=0pt,itemsep=0pt, parsep=0pt, topsep=0pt]
\item First framework using LLMs to dynamically optimize augmentation policies, eliminating the need for manual design and search-based approaches.
\item The adaptive method refines policies during the training loop, reducing the compute costs significantly by eliminating the need for multiple training iterations.
\item Domain-specific adaptation demonstrates consistent gains on medical imaging tasks over traditional methods.
\item Provides human-readable justifications for augmentation choices, enhancing transparency.
\end{itemize}
    


\section{RELATED WORK}

Early augmentation methods use combinations of basic
transformations like flipping, rotating, and cropping to manipulate an image to mitigate overfitting. However, finding the optimal combination of augmentation techniques for a specific task is challenging. To address this issue, several automated augmentation techniques have been developed. AutoAugment \cite{cubuk2019autoaugment} optimizes the augmentation policy through a reinforcement-learning-based approach, where the augmentation policy is selected from a search space in every iteration. RandAugment \cite{cubuk2019randaugment} and TrivialAugment \cite{trivialaugment} use randomness to select from a predefined set of augmentation techniques. In contrast, AugMix \cite{augmix} mixes differently augmented images to improve robustness.

Recently, Large Language Models (LLMs) have started to be used for decision-making problems. Studies such as \cite{eval1} and \cite{eval2} demonstrate the use of LLMs in evolutionary algorithm optimization. LLMs have also been applied to a range of decision-making tasks in AI. In \cite{zhang2023llm}, ChatGPT is utilized to tune fundamental hyperparameters such as batch size, learning rate, and optimizer type iteratively to improve model training. LLMs are also used for neural architecture search as a black-box optimizer \cite{nas1}, \cite{nas2}. In \cite{automl}, a fully LLM-driven training pipeline idea is presented. These recent studies highlight how LLMs are becoming integral to the AI model training process as decision-makers through their broad and contextual understanding of various training challenges.

LLMs that can generate textual data are
also used in generative data augmentation. In \cite{gen1}, \cite{gen2}, \cite{gen3}
it has been shown that these models can be used to create
synthetic data to enhance the diversity and size of existing
datasets. These approaches augment by generating new
examples using related embeddings, while our work focuses on
obtaining optimal augmentation policies without adding any
new examples to existing datasets.  

\section{Methodology}
\subsection{Overview of the Approach}

The overall system consists of the following core components: (1) an LLM responsible for suggesting augmentation policies based on dataset characteristics and prior model performance, (2) a deep learning model trained using the recommended augmentation policies, and (3) a feedback mechanism that reports performance back to the LLM for policy refinement. We propose two methods built on this framework, differing mainly in the timing and frequency of policy updates—factors that significantly affect both computational efficiency and generalization.

Let $\mathcal{A} = \{a_1, a_2, \dots, a_K\}$ denote the set of all possible augmentation operations. Each operation $a_k \in \mathcal{A}$ is parameterized by a set of values $\theta_k \in \Theta_k$, where $\Theta_k$ defines the valid range of parameters for the operation $a_k$ (e.g., rotation angle, blur radius). An individual augmentation operation can therefore be represented as a tuple $(a_k, \theta_k)$, where $\theta_k \in \Theta_k$.
An \textbf{augmentation policy} $\pi$ is defined as a set of $N$ augmentation operations:
\[
\pi = \left\{ (a_{k_1}, \theta_{k_1}), \dots, (a_{k_N}, \theta_{k_N}) \right\}
\]
Each policy is applied sequentially to input samples during training.
The space of all possible augmentation policies is given by:
\[
\Pi = \left\{ \pi \subseteq (\mathcal{A} \times \Theta) \mid |\pi| = N \right\}
\]

where $\Theta$ is the full augmentation parameter space. The LLM selects augmentation policies $\pi \in \Pi$ based on dataset characteristics and model feedback.

\subsubsection{Method 1: LLM-Guided Augmentation Policy Optimization}

\begin{algorithm}
\caption{LLM-Guided Augmentation Policy Optimization (Method 1)}
\label{alg:method1}
\begin{algorithmic}[1]
\Require Dataset $\mathcal{D}$, model architecture $\mathcal{M}$, max iterations $T$, number of augmentations $N$, initial prompt $P_0$ describing dataset, model architecture, performance goal

\State Initial augmentation policy: $\pi_0 \gets \texttt{LLMAPI:Query}(P_0)$
\State Initialize model: $\mathcal{M}_0 \gets \texttt{InitializeModel}(\mathcal{M})$

\For{$t = 1$ to $T$}
    \State Augment training data: $\mathcal{D}^{\text{aug}} \gets \texttt{ApplyPolicy}(\mathcal{D}_{\text{train}}, \pi_{t-1})$
    \State Train model on augmented data: $\mathcal{M}_t.\texttt{Train}(\mathcal{D}^{\text{aug}})$
    \State Evaluate on validation set: $A_t \gets \mathcal{M}_t.\texttt{Evaluate}(\mathcal{D}_{\text{valid}})$
    \State Construct feedback prompt: $P_t \gets \texttt{LLMAPI:SelfConstructFeedbackPrompt}(\pi_{t-1}, A_t, \mathcal{C})$
    \State Update augmentation policy: $\pi_t \gets \texttt{LLMAPI:Query}(P_t)$
\EndFor

\State \Return $\mathcal{M}_T$
\end{algorithmic}
\end{algorithm}

 This strategy follows an iterative process where augmentation policies are refined based on validation performance from previous training runs (Algorithm~\ref{alg:method1}). Initially, the LLM receives a prompt containing the dataset description, model architecture, performance goals, and the number of augmentation types to be used. It then generates a policy based on its prior knowledge and the provided context. The model is trained using this policy, and validation accuracy is computed and sent back to the LLM. Using this feedback, the LLM updates the policy each iteration, repeating for a set number of cycles to progressively refine the augmentation strategy.

\subsubsection{Method 2: Adaptive LLM-Guided Augmentation Policy Optimization}
The alternative dynamic in-training method refines augmentation policies continuously during training (Algorithm~\ref{alg:method2}). The LLM initially receives a prompt with dataset details, model architecture, performance goals, and the number of augmentation types to be used. During training, at every \textit{T} epochs, validation metrics and the current augmentation policy are sent to the LLM. It processes this feedback and returns an updated policy tailored to the model’s state which is applied immediately, allowing training to continue seamlessly. This iterative refinement continues until training is complete, enabling adaptive augmentation based on real-time performance.

\begin{algorithm}
\caption{Adaptive LLM-Guided Augmentation Policy Optimization (Method 2)}
\label{alg:method2}
\begin{algorithmic}[1]
\Require Dataset $\mathcal{D}$, model architecture $\mathcal{M}$, total epochs $E$, update interval $T$, number of augmentations $N$, initial prompt $P_0$ describing dataset, model architecture, performance goal

\State Initial augmentation policy: $\pi \gets \texttt{LLMAPI:Query}(P_0)$
\State Initialize model: $\mathcal{M}_0 \gets \texttt{InitializeModel}(\mathcal{M})$

\For{$e = 1$ to $E$}
    \State Apply augmentation policy: $\mathcal{D}^{\text{aug}} \gets \texttt{ApplyPolicy}(\mathcal{D}_{\text{train}}, \pi)$
    \State Train one epoch: $\mathcal{M}_{e} \gets \mathcal{M}_{e-1}.\texttt{TrainEpoch}(\mathcal{D}^{\text{aug}})$

    \If{$e \bmod T = 0$}
        \State Evaluate model: $A_e \gets \mathcal{M}_{e}.\texttt{Evaluate}(\mathcal{D}_{\text{valid}})$
        \State Construct feedback prompt: $P_e \gets \texttt{LLMAPI:SelfConstructFeedbackPrompt}(\pi, A_e, \mathcal{C})$
        \State Update policy: $\pi \gets \texttt{LLMAPI:Query}(P_e)$
    \EndIf
\EndFor

\State \Return $\mathcal{M}_E$
\end{algorithmic}
\end{algorithm}









This adaptive approach offers several advantages. It enables real-time evolution of augmentation policies, allowing modifications tailored to the model’s learning stage and dataset characteristics. By adapting during training, the method improves generalization and reduces the risk of overfitting. It also promotes broader exploration by exposing the model to a wider range of transformations across epochs. Unlike the before-training strategy, this method avoids repeated full retraining cycles. Instead, it operates with a single training session where policies are updated every $T$ epochs, significantly lowering computational cost.





\subsubsection{Model Architecture and Setup}
Augmentation policy selection is performed via pretrained LLMs such as ChatGPT-4o\footnote{\url{https://chat.openai.com/}}, Gemini 2.0\footnote{\url{https://gemini.google.com/}}, and DeepSeek\footnote{\url{https://deepseek.ai/}}, all accessed through an API. Each query includes dataset details, model specs, and prior validation results. The LLM uses this context to generate augmentation policies by drawing on its contextual knowledge of transformation effects and dataset characteristics. Particularly in medical imaging, where factors like contrast, anatomical structures, and artifacts matter, the LLM suggests domain-aware augmentations that enhance robustness. The API-based design supports scalable, real-time policy refinement without any manual input.

The LLM selects augmentations dynamically from the full \texttt{torchvision.transforms} library \footnote{\url{https://github.com/pytorch/vision}}, without restrictions. Deterministic prompting (temperature = 0) ensures reproducibility. This flexibility allows adaptation to dataset characteristics and model feedback. Models were trained using PyTorch and TensorFlow for 100 epochs with early stopping (patience = 10), using fixed hyperparameters for learning rate, batch size, and optimizer.

As the structure and phrasing of prompts played a critical role in the quality and effectiveness of the resulting augmentation policies, a detailed analysis is provided in Appendix \ref{appendixA}.

\section{Experimental Setup and Datasets}
We evaluated our LLM-guided augmentation methods against traditional strategies—RandAugment, TrivialAugment, and AugMix. TrivialAugment applies one random transformation per image, while AugMix uses three augmentation chains with up to three operations each. For RandAugment and LLM-based methods, we report results for N=2 and N=3 augmentations. This comparison highlights the effectiveness of real-time, dynamically refined augmentation policies.

The experimental setup included four diverse medical imaging datasets requiring specialized augmentations: (1) \textit{APTOS 2019 Blindness Detection} \cite{aptos2019}, with 3,662 retinal fundus images labeled into five diabetic retinopathy stages following ICDRSS; (2) \textit{Melanoma Cancer Image Dataset} \cite{melanoma}, comprising 13,900 $224\times224$ images of benign and malignant skin lesions; (3) \textit{Alzheimer-Parkinson MRI Dataset} \cite{alzheimer}, including RGB brain scans ($176\times208$) for healthy, AD, and PD classification; and (4) \textit{LIMUC Colonoscopy Dataset} \cite{limuc1,limuc3}, with 11,276 images labeled using the Mayo Endoscopic Score.
We used several standard deep architectures —ResNet-18 \cite{resnet}, MobileNetV2 \cite{mobilenet}, DenseNet \cite{densenet}, and ViT \cite{vit}—to ensure generalizability.
\section{Results and Discussion}
Table~\ref{tab1} shows that  LLM-driven augmentation consistently achieves the best results on APTOS2019 and has robust performance across all model architectures. While TrivialAugment and AugMix outperform no augmentation, their gains are modest and vary by architecture. RandAugment yields inconsistent results across models, suggesting that fixed augmentation magnitudes may not generalize well across different models.
Adaptive LLM-Guided Optimization (N=3, updated every epoch) delivers the highest gains, highlighting the value of frequent feedback-driven updates. Even with updates every five epochs, performance remains strong offering a balance between accuracy and computational efficiency.

Table~\ref{tab:merged_results} also shows that traditional augmentation methods perform poorly on the Melanoma dataset. TrivialAugment and AugMix even fall below the no-augmentation baseline. RandAugment also yields inconsistent results across models. These outcomes suggest that predefined augmentation policies may distort subtle visual cues essential for accurate classification in medical imaging.

On the other hand, the proposed LLM-Guided Augmentation further improves upon the baseline and standard methods. Notably, Adaptive method achieves the highest accuracy gains. Updating augmentations every epoch yields the best results, while updating every five epochs still maintains a substantial advantage. These results confirm that dynamic augmentation enhances model robustness in high-variance medical imaging datasets.



\begin{table}[ht]
\caption{Validation accuracy comparison of augmentation strategies on APTOS2019 and Melanoma datasets.}
\label{tab:merged_results}
\centering
\begin{tabular}{|l|c|c|c|c|c|c|}
\hline
\textbf{Augmentation Method} & \multicolumn{3}{c|}{\textbf{APTOS2019}} & \multicolumn{3}{c|}{\textbf{Melanoma}} \\
\cline{2-7}
                             & ResNet18 & MobileNetV2 & ViT32 & ResNet18 & MobileNetV2 & ViT32 \\
\hline
No Augmentation & 0.8388 & 0.8415 & 0.7978 & 0.9061 & 0.8998 & 0.8161 \\
TrivialAugment  & 0.8497 & 0.8706 & 0.8051 & 0.8838 & 0.8910 & 0.7963 \\
AugMix          & 0.8525 & 0.8469 & 0.8005 & 0.8864 & 0.8981 & 0.7471 \\
\hline
\multicolumn{7}{|c|}{\textbf{N=2 (Two Augmentations Allowed)}} \\
\hline
RandAugment                  & 0.8607 & 0.8388 & 0.8197 & 0.8902 & 0.8973 & 0.7627 \\
LLM-Guided (ChatGPT)         & 0.8689 & 0.8716 & 0.8212 & 0.9078 & 0.9125 & 0.8352 \\
LLM-Guided (Gemini)          & 0.8743 & 0.8661 & 0.8205 & 0.9087 & 0.9028 & 0.8244 \\
LLM-Guided (DeepSeek)        & 0.8770 & 0.8777 & 0.8226 & 0.9091 & 0.9091 & 0.8401 \\
LLM-Guided (Adaptive, @1 Ep) & 0.8743 & 0.8743 & 0.8319 & \textbf{0.9506} & \textbf{0.9352} & \textbf{0.8512} \\
LLM-Guided (Adaptive, @5 Ep) & 0.8743 & 0.8743 & 0.8305 & 0.9468 & 0.9336 & 0.8433 \\
\hline
\multicolumn{7}{|c|}{\textbf{N=3 (Three Augmentations Allowed)}} \\
\hline
RandAugment                  & 0.8470 & 0.8607 & 0.8169 & 0.8918 & 0.8960 & 0.7726 \\
LLM-Guided (ChatGPT)         & 0.8743 & 0.8716 & 0.8240 & 0.9070 & 0.9078 & 0.8198 \\
LLM-Guided (Gemini)          & 0.8661 & 0.8743 & 0.8219 & 0.8965 & 0.8897 & 0.8202 \\
LLM-Guided (DeepSeek)        & 0.8675 & 0.8750 & 0.8275 & 0.9529 & 0.9367 & 0.8310 \\
LLM-Guided (Adaptive, @1 Ep) & \textbf{0.8798} & \textbf{0.8798} & \textbf{0.8368} & \textbf{0.9738} & \textbf{0.9668} & 0.8417 \\
LLM-Guided (Adaptive, @5 Ep) & 0.8756 & 0.8756 & 0.8771 & 0.9506 & 0.9344 & 0.8456 \\
\hline
\end{tabular}
\label{tab1}
\end{table}
Table~\ref{tab2} shows that while traditional methods underperformed on the Melanoma dataset, RandAugment (N=2) shows strong results on the Alzheimer-Parkinson Dataset, suggesting that it may be a viable option in some contexts. In contrast, TrivialAugment and AugMix perform poorly, likely due to their transformations not aligning well with the dataset’s structure. LLM-Guided methods performs strongly, with ChatGPT (N=2) outperforming all baselines, with Gemini performing slightly behind but better than traditional methods. DeepSeek achieves the highest accuracy, surpassing even the adaptive method on ResNet18 and MobileNetV2. Although the adaptive method remains competitive, it lags behind the non-adaptive method on this dataset.

\begin{table}[ht]
\caption{Validation accuracy comparison of augmentation strategies on Alzheimer-Parkinson and LIMUC datasets.}
\label{tab:merged_alz_limuc}
\centering
\begin{tabular}{|l|c|c|c|c|c|c|}
\hline
\textbf{Augmentation Method} & \multicolumn{3}{c|}{\textbf{Alzheimer-Parkinson}} & \multicolumn{3}{c|}{\textbf{LIMUC}} \\
\cline{2-7}
                             & ResNet18 & MobileNetV2 & ViT32 & ResNet18 & DenseNet121 & ViT32 \\
\hline
No Augmentation              & 0.9422 & 0.9037 & 0.7809 & 0.7599 & 0.7648 & 0.6910 \\
TrivialAugment               & 0.8981 & 0.8629 & 0.7787 & 0.7660 & 0.7660 & 0.7008 \\
AugMix                       & 0.9267 & 0.9352 & 0.7815 & 0.7413 & 0.7413 & 0.7018 \\
\hline
\multicolumn{7}{|c|}{\textbf{N=2 (Two Augmentations Allowed)}} \\
\hline
RandAugment                  & 0.9684 & 0.9444 & 0.8234 & 0.7636 & 0.7512 & 0.7172 \\
LLM-Guided (ChatGPT)         & 0.9684 & 0.9614 & 0.8454 & 0.7784 & 0.7673 & 0.7373 \\
LLM-Guided (Gemini)          & 0.9622 & 0.9534 & 0.8333 & 0.7611 & 0.7748 & 0.7359 \\
LLM-Guided (DeepSeek)        & \textbf{0.9738} & 0.9668 & 0.8404 & 0.7611 & 0.7748 & 0.7373 \\
LLM-Guided (Adaptive, @1 Ep) & 0.9526 & 0.9460 & 0.8666 & 0.7883 & 0.7772 & 0.7491 \\
LLM-Guided (Adaptive, @5 Ep) & 0.9468 & 0.9452 & 0.8660 & 0.7676 & 0.7712 & 0.7491 \\
\hline
\multicolumn{7}{|c|}{\textbf{N=3 (Three Augmentations Allowed)}} \\
\hline
RandAugment                  & 0.9630 & 0.9174 & 0.8311 & 0.7561 & 0.7660 & 0.7265 \\
LLM-Guided (ChatGPT)         & 0.9676 & 0.9483 & 0.8354 & 0.7748 & 0.7847 & 0.7454 \\
LLM-Guided (Gemini)          & 0.9614 & 0.9550 & 0.8391 & 0.7587 & 0.7834 & 0.7468 \\
LLM-Guided (DeepSeek)        & 0.9707 & \textbf{0.9707} & 0.8660 & 0.7611 & 0.7748 & 0.7447 \\
LLM-Guided (Adaptive, @1 Ep) & 0.9701 & 0.9405 & \textbf{0.8728} & \textbf{0.7919} & \textbf{0.7852} & \textbf{0.7605} \\
LLM-Guided (Adaptive, @5 Ep) & 0.9684 & 0.9475 & 0.8712 & 0.7892 & 0.7834 & 0.7576 \\
\hline
\end{tabular}
\label{tab2}
\end{table}

Table~\ref{tab2} also displays the results on the LIMUC dataset. On this dataset, traditional automated augmentation methods provide only modest improvements or even degrade performance compared to the no augmentation baseline. On the other hand, LLM-Guided methods yield significant gains, with the best results achieved by our Adaptive method (N=3) using updates at every epoch. 

\subsection{LLM Reasoning and Chain-of-Thought Behavior}
While generating augmentation policies, LLMs consistently demonstrated structured, task-aware chain-of-thought reasoning informed by data traits and model performance, rather than relying on random or fixed rules. They regularly referenced dataset-specific characteristics—avoiding color augmentations on grayscale brain MRIs in favor of spatial transformations like rotation and blurring, and preserving anatomical features in LIMUC by choosing brightness or contrast changes to simulate lighting variability.
LLMs also adapted augmentation strength based on validation feedback: they hardened transformations when accuracy improved and recommended softer ones when performance declined. One response noted: \textit{“The validation accuracy dropped slightly from 0.9514 to 0.9468 in the second iteration, indicating that the updated policy may have introduced too much variability or distortion, which could be counterproductive for the model's performance. To address this, we will refine the augmentation policy by reducing the intensity of some augmentations and introducing new ones that are less likely to distort critical features in the brain MRI images.”}

Beyond performance tuning, LLMs provided domain-specific explanations for individual augmentations. For example, on the Alzheimer dataset, one rationale stated: \textit{“GaussianBlur simulates slight blurring, which can occur in medical imaging due to motion or noise. This augmentation helps the model become more robust to minor image quality variations.”} These cause-and-effect explanations reflect semantic awareness and introspective reasoning capabilities of LLMs, absent in traditional augmentation search methods. Among the models tested, DeepSeek gave the most coherent and domain-relevant justifications, while Gemini and ChatGPT sometimes used more generic language, though all produced strong policies.

These observations reinforce the idea that LLMs can act as intelligent decision-makers in data-centric workflows. Their ability to rationalize policy choices and adapt based on domain feedback provides a unique advantage over conventional augmentation search approaches.

\subsection{Time Complexity Analysis}
\textit{TrivialAugment}, \textit{RandAugment}, and \textit{AugMix} use fixed or randomized policies without dataset-specific optimization, incurring no additional overhead and a time complexity of $\mathcal{O}(T)$, where $T$ is the time to train a model. In contrast, search-based methods like \textit{AutoAugment} and its successors rely on reinforcement learning or density matching over multiple trials, requiring $K$ full training runs and resulting in $\mathcal{O}(K \cdot T)$ time complexity, with $K \gg 1$. This makes them impractical for domains like medical imaging. Our LLM-guided (before-training) strategy reduces this to $\mathcal{O}(N \cdot T)$, where $N \ll K$, as prior knowledge is used to converge with fewer iterations. The adaptive approach further reduces complexity to $\mathcal{O}(T)$ by requiring only a single training run with lightweight LLM queries during training. These queries are negligible in cost, typically taking less than one second and costing approximately \$0.003 per iteration, making the adaptive method both efficient and accurate, matching the speed of heuristic methods while outperforming them.

\section{Conclusion}
In this work, we introduced two LLM-driven strategies for optimizing data augmentation policies based on model performance feedback. By harnessing the contextual understanding and reasoning capabilities of large language models, our methods eliminate the need for manual tuning or expensive search-based optimization, while outperforming traditional approaches. Experiments on four diverse medical imaging datasets demonstrate that the proposed methods consistently improve model accuracy, particularly in domains where subtle visual patterns are critical. Notably, the adaptive approach, which updates augmentation policies in-training based on intermediate feedback, achieves the highest accuracy while significantly reducing computational overhead compared to full retraining cycles.

Beyond data augmentation, the results imply that LLMs can play an active role in optimizing various components of machine learning workflows, from hyperparameter tuning to architecture search. The ability of LLMs to interpret dataset characteristics and training signals positions them as powerful tools for automating and adapting training strategies in complex, domain-specific tasks. Future work will explore integrating vision-language models (VLMs) to enhance semantic understanding in augmentation decisions, potentially improving performance in multimodal or context-sensitive applications. We also aim to investigate the interpretability and robustness of LLM-generated policies and  extend our framework to other domains such as satellite imagery and industrial inspection to evaluate its generalizability and scalability.



\appendix    
\acknowledgments 
 
This work has been supported by Middle East Technical University Scientific Research Projects
Coordination Unit under grant number ADEP-704-2024-11486. 

\section{Prompt Analysis and Ablation Studies} 
\label{appendixA}
The prompts were iteratively refined in response to specific failure cases. Below, we outline the key prompt design challenges we encountered and the strategies we employed to address them. These refinements were applied consistently across all experiments and significantly enhanced the stability, diversity, and contextual appropriateness of LLM-generated policies.

\begin{itemize}[left=0pt,itemsep=0pt, topsep=0pt] 
\item{\textbf{Avoidance of Technique Changes Across Iterations:}}  
In early experiments, the LLM often reused the same augmentation types, modifying only parameters like rotation degree or probability. This conservative behavior, likely due to interpreting prior policies as constraints, led to limited diversity and occasional overfitting. To address this, we added an explicit instruction in the after-iteration prompt:
\textit{“You are free to remove or add new augmentation techniques, as long as the total number remains the same.”}
Following this change, the LLM began proposing more varied transformations, improving performance on complex or noisy datasets.

\item{\textbf{Adverse Effects of Manually Constraining the Augmentation Space:}}  
Initially, we attempted to stabilize policy generation by limiting the augmentation options in the prompt to a small predefined set. However, this led to repetitive and underperforming policies that failed to capture dataset-specific variation. Once we allowed the LLM to select freely from the full torchvision.transforms space, both diversity and performance improved across datasets—highlighting the value of open-ended selection guided by context.

\item{\textbf{Loss of Context in Iterative Feedback:}}  
We observed that in follow-up iterations, the LLM sometimes ignored dataset and model details from the initial prompt, defaulting to generic augmentations (e.g., color jitter on grayscale images or excessive flips on asymmetric structures). To mitigate this, we added a reminder in each feedback prompt: \textit{“Always tailor your augmentation suggestions to the specific properties of the dataset and model described earlier.”} This improved policy relevance, especially for datasets like APTOS and LIMUC.

\item{\textbf{Inability to Recover from Suboptimal Policies:}}  
Another issue arose when a poor policy caused a drop in validation accuracy and subsequent iterations made only minor tweaks instead of reverting to more effective earlier alternatives. We inferred that the LLM was optimizing to improve the most recent result, rather than seeking a globally optimal policy. To address this, we updated the prompt to state:  
\textit{“Your objective is to find the best augmentation policy overall, not just to improve the last one.”}  
This adjustment encouraged more exploratory behavior, reduced policy collapse, and helped the LLM recover from earlier mistakes.

\end{itemize}

\bibliography{report} 
\bibliographystyle{spiebib} 
\section{AUTHORS\textquoteright\ BACKGROUND}
\begin{table}[h]
\centering
\begin{tabular}{@{}l@{\hskip 1em}c@{}}
\begin{tabular}{|l|l|l|l|}
\hline
\textbf{Your Name} & \textbf{Title*} & \textbf{Research Field} & \textbf{Personal website} \\
\hline
Ant Duru& Master Student & LLM-Guided Training Workflows  &  \href{https://www.linkedin.com/in/ant-duru-295b3615a/}{LinkedIn}\\
\hline
Alptekin Temizel & Full Professor & Computer Vision, Deep Learning  & \url{https://blog.metu.edu.tr/atemizel/} \\
\hline
\end{tabular}
\end{tabular}
\caption{Authors' Background}
\end{table}

\end{document}